\documentclass[10pt, conference, compsocconf]{IEEEtran}

\usepackage{wrapfig}
\usepackage[pdftex]{graphicx}
\graphicspath{ {./Images/} }
\DeclareGraphicsExtensions{.pdf,.png}
\usepackage[export]{adjustbox}
\usepackage{multicol}
\usepackage{booktabs}
\usepackage{tabu}
\usepackage{multirow}

\ifCLASSINFOpdf
\else
\fi
\begin{document}
%
% paper title
% can use linebreaks \\ within to get better formatting as desired
\title{Anomaly Detection in Images}

% author names and affiliations
% use a multiple column layout for up to two different
% affiliations

\author{\IEEEauthorblockN{Manpreet Singh Minhas, John Zelek }
\IEEEauthorblockA{Systems Design Engineering\\
University of Waterloo\\
Waterloo, Canada\\
Email: \{msminhas,jzelek\}@uwaterloo.ca}

}

% conference papers do not typically use \thanks and this command
% is locked out in conference mode. If really needed, such as for
% the acknowledgment of grants, issue a \IEEEoverridecommandlockouts
% after \documentclass

% for over three affiliations, or if they all won't fit within the width
% of the page, use this alternative format:
% 
%\author{\IEEEauthorblockN{Michael Shell\IEEEauthorrefmark{1},
%Homer Simpson\IEEEauthorrefmark{2},
%James Kirk\IEEEauthorrefmark{3}, 
%Montgomery Scott\IEEEauthorrefmark{3} and
%Eldon Tyrell\IEEEauthorrefmark{4}}
%\IEEEauthorblockA{\IEEEauthorrefmark{1}School of Electrical and Computer Engineering\\
%Georgia Institute of Technology,
%Atlanta, Georgia 30332--0250\\ Email: see http://www.michaelshell.org/contact.html}
%\IEEEauthorblockA{\IEEEauthorrefmark{2}Twentieth Century Fox, Springfield, USA\\
%Email: homer@thesimpsons.com}
%\IEEEauthorblockA{\IEEEauthorrefmark{3}Starfleet Academy, San Francisco, California 96678-2391\\
%Telephone: (800) 555--1212, Fax: (888) 555--1212}
%\IEEEauthorblockA{\IEEEauthorrefmark{4}Tyrell Inc., 123 Replicant Street, Los Angeles, California 90210--4321}}

% use for special paper notices
%\IEEEspecialpapernotice{(Invited Paper)}

% make the title area
\maketitle

\begin{abstract}
Visual defect assessment is a form of anomaly detection.  This is very relevant in finding faults such as cracks and markings in various surface inspection tasks like pavement and automotive parts. The task involves detection of deviation/divergence of anomalous samples from the normal ones. Two of the major challenges in supervised anomaly detection are the lack of labelled training data and the  low availability of anomaly instances. Semi-supervised methods which learn the underlying distribution of the normal samples and then measure the deviation/divergence from the estimated model as the anomaly score have limitations in their overall ability to detect anomalies. This paper proposes the application of network-based deep transfer learning using convolutional neural networks (CNNs) for the task of anomaly detection. Single class SVMs have been used in the past with some success, however we hypothesize that deeper networks for single class classification should perform better. Results obtained on established anomaly detection benchmarks as well as on a real-world dataset, show that the proposed method clearly outperforms the existing state-of-the-art methods, by achieving a staggering average area under the receiver operating characteristic curve value of 0.99 for the tested datasets which is an average improvement of 41\% on the CIFAR10, 20\% on MNIST and 16\% on Cement Crack datasets. 

\end{abstract}


% For peer review papers, you can put extra information on the cover
% page as needed:
% \ifCLASSOPTIONpeerreview
% \begin{center} \bfseries EDICS Category: 3-BBND \end{center}
% \fi
%
% For peerreview papers, this IEEEtran command inserts a page break and
% creates the second title. It will be ignored for other modes.
\IEEEpeerreviewmaketitle

\section{Introduction}
% no \IEEEPARstart
Anomaly detection refers to the problem of finding patterns in data that do not conform to expected behavior. These non-conforming patterns are often referred to as anomalies, outliers, discordant observations, exceptions, aberrations, surprises, peculiarities or contaminants in diﬀerent application domains \cite{Chandola:2009:ADS:1541880.1541882}.  With the current proliferation of data, humongous volumes of both structured and unstructured data is available at our disposal. The reason why anomaly detection is important is because anomalies are salient and contain interesting information that is typically of interest  in a majority of application domains. 

Anomaly detection techniques have a broad spectrum of application areas such as video surveillance, credit card fraud detection, surface defect detection, medical diagnostics etc. The detection methods can be broadly classified into three categories namely: (1) supervised; (2) semi-supervised;  and (3) unsupervised. Truly unsupervised anomaly detection techniques are virtually unavailable for images. Although clustering, flow based or predictive modelling techniques fall under this category, they are difficult to use for anomaly identification. Semi-supervised techniques involve the use of generative models such as Autoencoders (AEs) \cite{Lu2018AnomalyDF}, Generative Adversarial Networks (GANs) \cite{anogan}, \cite{ganomaly} or statistical approaches \cite{mdialgo} \cite{noveltypimentel} to learn/estimate the density function of the underlying distribution of the normal data implicitly or explicitly. Then a measure of divergence/deviation from this distribution is used to calculate an anomaly score which outputs the anomalous instances based on an appropriate threshold. 

Supervised approaches involve labelled training  \cite{Gornitz:2013:TSA:2512538.2512545}. Suitable pattern recognition or classification techniques can be applied to the task of supervised anomaly detection since it essentially translates to a binary classification problem (also referred to a single class classifier where the one class is the normal class with no anomalies and the other class contains what we call anomalies). Two of the major challenges in anomaly detection are lack of labelled data and low anomaly instances. Deep learning and particularly Convolutional Neural Networks (CNNs) which are a class of artificial neural networks, have emerged as very powerful tools for computer vision applications especially for  classification tasks.  Training these networks requires huge volumes of data and often training from scratch turns out to be unfeasible. Transfer learning is a tool that overcomes these challenges. To the best of our knowledge no previous study has been conducted on the application of network-based deep transfer learning using CNNs to the task of anomaly detection in images. 

Our main contribution is the application of transfer learning using different CNN architectures to the task of anomaly detection in images. Results obtained on CIFAR10 \cite{cifar10}, MNIST \cite{mnist} and Concrete Crack \cite{cementdata} datasets show that this approach outperforms the state-of-the-art techniques for anomaly detection.

% You must have at least 2 lines in the paragraph with the drop letter
% (should never be an issue)

%\subsection{Subsection Heading Here}
%Subsection text here.
%\subsubsection{Subsubsection Heading Here}
%Subsubsection text here.

\section{Transfer Learning} \label{transferlearning}
Deep learning networks  have shown to perform well in a variety of tasks  with applications ranging from Computer Vision and Natural Language Processing to Speech recognition. However, to train these deep networks, very huge volumes of training data is required. And for most applications training a network from scratch is impracticable. This is because the collection of data is complex and expensive. Getting good quality annotations can be difficult to obtain due to the monotonous nature of the task.  This makes it extremely difficult to build a large-scale, high-quality annotated dataset \cite{tranferlearningsurvey}.  
Transfer Learning is a machine learning tool used to tackle this challenge. The goal of transfer learning is to improve learning in a target task by leveraging knowledge from a source task. \cite{Torrey2009TransferL}

In \cite{tranferlearningsurvey}, Chuanqi Tan et al. classify deep transfer learning into four categories: instances-based deep transfer learning, mapping-based deep transfer learning, network-based deep transfer learning, and adversarial based deep transfer learning. We use network-based deep transfer learning method. Network-based deep transfer learning refers to the reuse a partial network pre-trained in the source domain, including its network structure and connection parameters and transferring it to be a part of deep neural network which used in target domain \cite{tranferlearningsurvey}. In this type of transfer learning, the source network is thought of as consisting of two sub-networks:
\begin{enumerate}
    \item Feature extractor sub-network,
    \item Classification sub-network.
\end{enumerate}
The target network is constructed using the source network with some modifications and trained on the target dataset for the intended task. 

\section{Methodology} \label{sec:methodology}

The steps involved in the approach proposed for anomaly detection using network-based deep transfer learning are as follows:
\begin{enumerate}
    \item Source Model Selection:
    The first step is selection of a model architecture which is pre-trained for some source task on a huge dataset belonging to the source domain. The choice of the architecture depends on the anomaly detection task. The selected architecture and its pre-trained weights are used as a starting point for the target model for the anomaly detection task. 
    \item Target model training:
    The target model is now ready for training. There are two strategies that can be used here: \cite{cs231n}
    \begin{enumerate}
        \item We can use a CNN as a fixed feature extractor: The pre-trained network is taken and its last fully-connected layer is removed. The network then behaves as a fixed feature extractor. Subsequently, we train a softmax classifier for the target dataset with the fixed features as input. Note that the earlier layers are frozen,  i.e., the weights are not changed during training of the classifier.
        \item The resulting network is Fine Tuned: If the fixed feature extractor approach gives one inadequate results then proceed to fine tuning. In addition to the softmax classifier, few layers of the pre-trained network are unfrozen during  backpropagation. In deep neural networks (DNNs), the hidden layers can be considered as increasingly complex feature transformations and the final softmax layer as a log-linear classifier making use of the most abstract features computed in the hidden layers \cite{cltransfer}. Therefore, the earlier layers are kept frozen during fine tuning. This is because they are good at extracting generic features useful in other tasks as well. Also, the learning rate is set lower than normal training. This is because the weights learned are good and we don't want to change the weights too fast and too much. 
    \end{enumerate}
\end{enumerate}

\begin{figure*}
    \centering
    \includegraphics[width=14cm,frame]{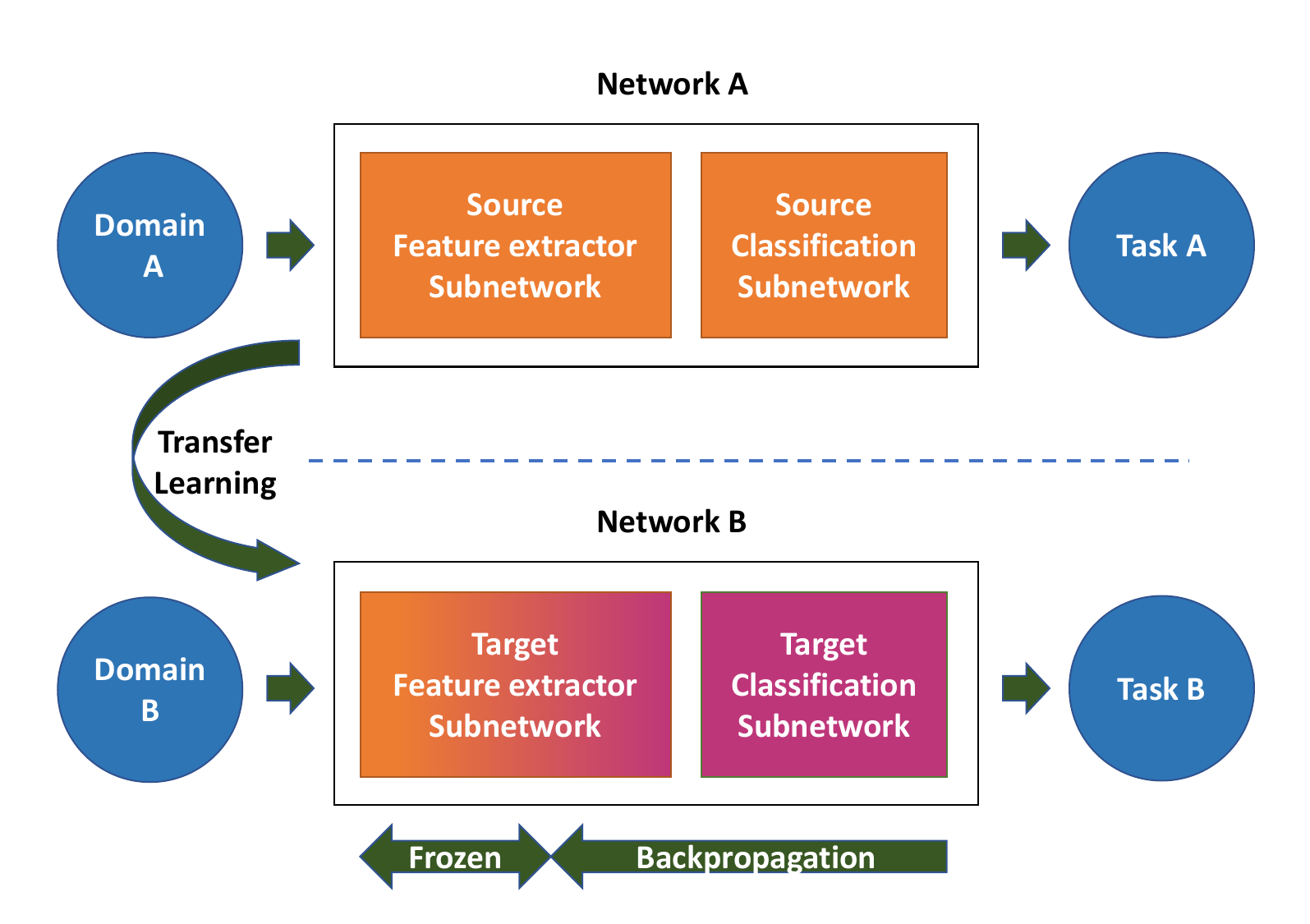}
    \caption{Illustration of a network based deep transfer learning from a source domain A and task A to target domain B and task B. The Network A is trained on a large training dataset and is called the pre-trained network. Network B is constructed by using parts of Network A followed by a new softmax classification network. Finally the resulting network B is initialized with the pre-trained weights and trained using backpropagation on the target dataset.}
    \label{fig:transferlearning}
\end{figure*}

Fig. \ref{fig:transferlearning} summarizes the approach used for building a CNN for the task of anomaly detection using network-based deep transfer learning. 

\section{Experimental Setup}
This section introduces the experimental setup in terms of the datasets, source CNN architectures, implementation and training details as well as the evaluation criteria. Also, the anomaly detection tasks are explained for every dataset. 

\subsection{Datasets}
\begin{figure*}
    \centering
    \includegraphics[width=17.2cm,frame]{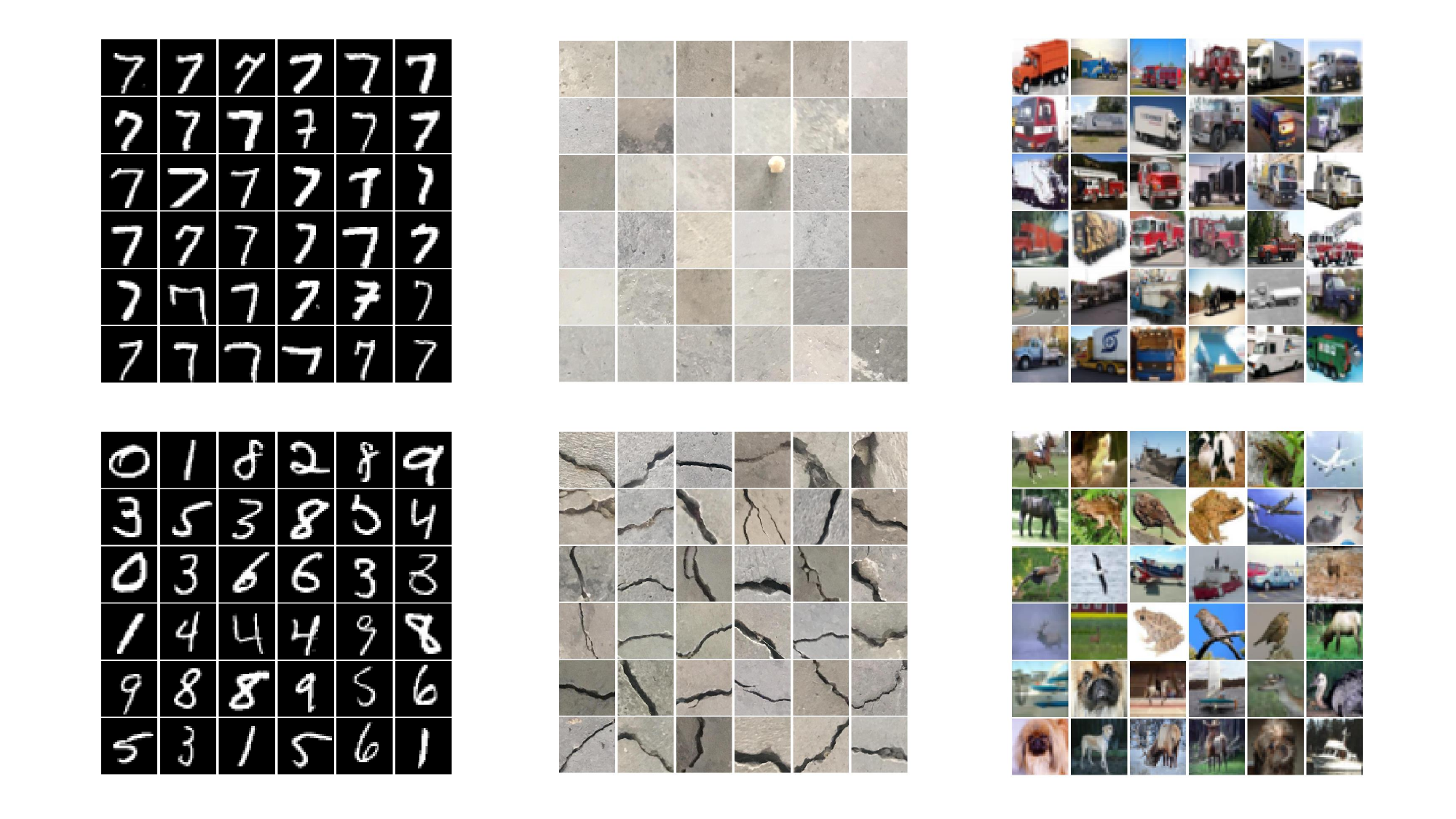}
    \caption{Exemplary images for the normal and anomaly classes for the MNIST \cite{mnist}, Cement Crack \cite{cementdata} and CIFAR10 \cite{cifar10} datasets from left to right.}
    \label{fig:exemplaryimages}
\end{figure*}

In order to evaluate the proposed method and demonstrate its performance on the task of anomaly detection, experiments were conducted on three different datasets namely CIFAR10, MNIST and Concrete Crack. For the experiments conducted on CIFAR10 and MNIST the anomaly detection task was one versus the rest approach. The anomaly class consisted of one class from these datsets and the normal class was constructed using random sampling from the remaining nine classes with size equal to the normal class. The fact that the samples were drawn randomly from the rest of the classes as a whole introduces slight class imbalance and makes the anomaly detection more challenging. Concrete cracks have well defined normal and anomaly classes which are negative and positive crack classes respectively. Exemplary images for normal and anomalous classes for all the three datasets are shown in Fig. \ref{fig:exemplaryimages}.

\subsubsection{CIFAR10} 
This dataset consists of 60,000 32x32 colour images in 10 classes, with 6,000 images per class. There are 50,000 training images and 10000 test images. Using the leave-one-out approach ten different combinations of anomaly and normal classes were constructed. For each combination there were 5k training samples per class and 1k test samples per class. 

\subsubsection{MNIST} 
This dataset consists of 60,000 28x28 grayscale images of the 10 digits, along with a test set of 10,000 images. For this dataset too ten different combinations of anomaly and normal classes were constructed using the leave-one-out approach. For each combination there were 6k training samples per class and 1k test samples per class. 

\subsubsection{Concrete Crack} 
The dataset contains concrete images with two classes namely positive and negative crack. There are 20,000 277x277 color images for each class. Experiments were conducted with 2000 training images per class and 4,000 test images per class. 

\subsection{CNN architectures}
Three state-of-the-art CNN architectures were selected for conducting experiments and are briefly described below. 

\subsubsection{DenseNet} Densely Connected Convolutional Networks \cite{Huang2017Densenet} (DenseNets) are the  latest addition to deep CNN architectures. Every layer is connected to every other layer in a feed forward fashion so that  the network with L layers has $\frac{L(L+1)}{2}$ direct connections. DenseNet-169 architecture is used as the source network for our experiments.

\subsubsection{ResNet} Deep Residual Networks \cite{He2016resnet} introduced the concept of identity shortcut connections that skip one or more layers. These were introduced in 2015 by Kaiming He. et.al. and bagged $1^{st}$ place in the ILSVRC 2015 classification competition . ResNet-152 architecture is used for the experiments.

\subsubsection{Inception} Inception architectures were introduced by Google as GoogLeNet / Inception-v1. The basic building block inception module passes the input from previous layers through multiple convolution layers and max pooling simultaneously and are concatenated together at output. This eliminates the need to think of which filter size to use at each layer. Inception-V4 \cite{Szegedy2017Inceptionv4IA} architecture is used for the experiments.

\subsection{Implementation}
Publicly available implementations in Keras \cite{chollet2015keras} of basic architecture and weights of these networks pre-trained on ImageNet were used as a starting point for all the three chosen architectures \cite{kerasapplicationgithub} \cite{githubinceptionv4}. As discussed in section \ref{sec:methodology}, after the model weights are loaded, the softmax layer is replaced with a new layer having two neurons for the anomaly detection task. Subsequently,  the modified network is fine tuned by backpropagation.

For all the three networks following training parameters are the same. Stochastic gradient descent optimizer is used with parameters values as learning rate = $10^{-3}$, decay = $10^{-6}$, momentum = 0.9 and nesterov = true. Batch size is 16 and shuffling is enabled. The model is trained for 50 epochs and the best model is used for the results. 
\begin{itemize}
    \item DenseNet-169: The input images are resized to (224,224) before feeding the model. 
    \item ResNet-152: The input images are resized to (224,224) before feeding the model.
    \item Inception-V4: The input images are resized to (299,299) before feeding the model.
\end{itemize}

The prediction probabilities from the normal class neuron are used as anomaly scores and for performing the anomaly detection evaluation.

For MNIST dataset RGB image is made by copying the greyscale values three times before re-sizing is performed.

\subsection{Evaluation}
Evaluation is done using the area under curve (AUC) measurement of the receiver operating characteristics (ROC) \cite{aucling}. The ROC curve is plotted with true positive rates (TPR) against the false positive rates (FPR). ROC is a probability curve and AUC represents degree or measure of separability. An excellent model has AUC near to the 1 which means it has good measure of separability. When AUC is 0.5, it means the model has no class separation capacity whatsoever.

\section{Results and Discussion}

AUC values of the experiments conducted for CIFAR10 and MNIST datasets  are summarized in Table \ref{tab:cifar10auc} and \ref{tab:mnistauc} respectively. The proposed method for all three chosen architectures clearly outperforms the previous work for the anomaly detection task across all the classes for both datasets. For the CIFAR10 dataset, the minimum, maximum and average improvement is 31\%, 58\% and 41\%. Even for the deer class for which all the other methods perform poorly, an AUC value of 0.99 is obtained. For the MNIST dataset, an average increase of 20\% is achieved over the other methods. 

Table \ref{tab:concretecrack} summarizes the results on the cement crack which is a real world dataset. The GANomaly \cite{ganomaly} model had to be trained on 16,000 samples to be able to achieve an AUC value of 0.858. Although for our transfer learning based approach all the three architectures were trained on only 2,000 images per class, the proposed method clearly outperforms by achieving an AUC value of 0.99. 

\begin{figure*}
    \centering
    \includegraphics[width=17.2cm,frame]{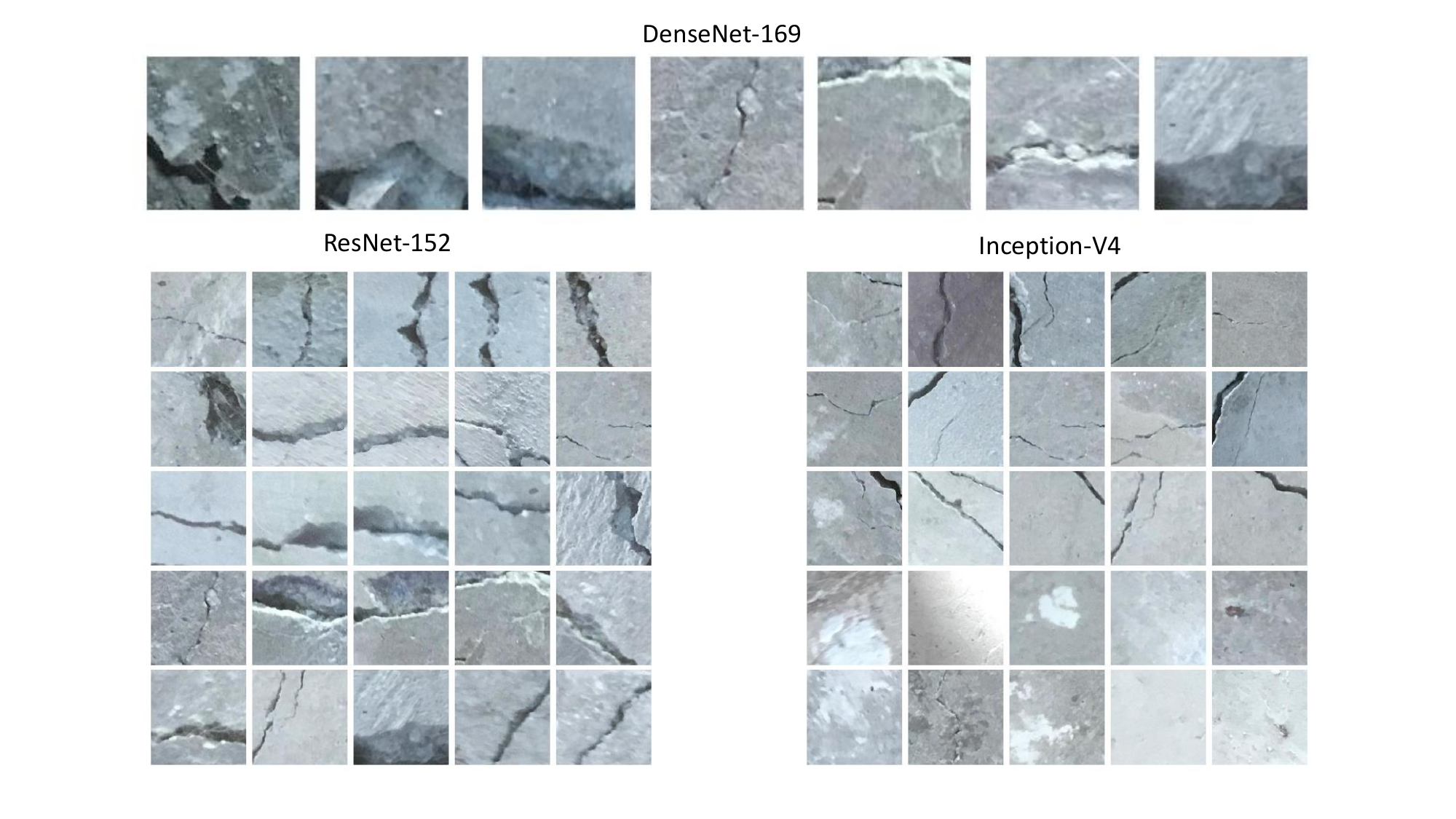}
    \caption{Few examples of miss-classified images for the Concrete Crack dataset for all the three architectures.}
    \label{fig:missclassifiedimages}
\end{figure*}

The confusion matrices for the cement crack experiments are shown in Tables \ref{tab:cmatrixdensenet}, \ref{tab:cmatrixresnet} and \ref{tab:cmatrixinceptionv4}. The architectures on an average have precision, recall and f1-score of 0.99, which indicates that the classifier is performing extremely well for the task of anomaly detection. Fig. \ref{fig:missclassifiedimages} shows the examples that were miss-classified in the experiments. It is important to note here that in order to show the capability of the proposed approach despite fewer training examples, the architectures were trained only on 2,000 images per class. The results are  better if more training data is available.

We see that even though all the three architectures achieve state-of-the-art results, there is variation among them. DenseNet-169 performs the best followed by ResNet-152 and finally Inception-V4. Even though on an average Inception-V4 performs lowest among the three architectures on the CIFAR10 dataset, it outperforms the other two on MNIST dataset. In the preliminary experiments conducted on the challenging German Asphalt Pavement Distress Dataset \cite{gapsdataset}, DenseNet-169 achieved an average F1 score of 95\% and AUC value of 0.91. 

The stellar results achieved despite the low number of training examples using the challenging AUC evaluation metric on benchmark as well as real world datasets indicate that the proposed method is highly suitable for anomaly detection.  The results also shown that it clearly outperforms the previous state-of-the-art methods.

\begin{table*}[!ht]
\centering
\caption{AUC results for CIFAR10 dataset}
\label{tab:cifar10auc}
\begin{tabular}{lllllllllll}
\toprule 
             & \multicolumn{10}{c}{CIFAR10}                                                                                                                                                                         \\ \cmidrule{2-11}
Model        & plane             & car               & bird              & cat               & deer              & frog              & horse             & ship              & truck             & dog               \\ \midrule
GANomaly \cite{ganomaly}    & 0.633             & 0.631             & 0.51              & 0.587             & 0.593             & 0.683             & 0.605             & 0.616             & 0.617             & 0.628             \\
AnoGAN \cite{anogan}      & 0.516             & 0.492             & 0.411             & 0.399             & 0.335             & 0.321             & 0.399             & 0.567             & 0.511             & 0.393             \\
EGBAD \cite{egbad}       & 0.577             & 0.514             & 0.383             & 0.448             & 0.374             & 0.353             & 0.526             & 0.413             & 0.555             & 0.481             \\
DenseNet-169 & \textbf{0.998449} & \textbf{0.998933} & 0.994980          & \textbf{0.992014} & 0.998145          & \textbf{0.991758} & 0.999031          & 0.998386          & \textbf{0.998948} & \textbf{0.998291} \\
ResNet-152   & 0.998071          & 0.998203          & \textbf{0.995249} & 0.991605          & \textbf{0.998480} & 0.991375          & \textbf{0.999607} & \textbf{0.999289} & 0.998934          & 0.997900          \\
Inception-V4 & 0.930263          & 0.971474          & 0.842340          & 0.853591          & 0.895042          & 0.893674          & 0.949273          & 0.921899          & 0.954804          & 0.931945          \\
\bottomrule
\end{tabular}
\end{table*}

\begin{table*}[!ht]
\centering
\caption{AUC results for MNIST dataset}
\label{tab:mnistauc}
\begin{tabular}{lllllllllll}
\toprule 
             & \multicolumn{10}{c}{MNIST}                                                                                                                                                                         \\ \cmidrule{2-11}
Model        & Class 0           & Class 1           & Class 2           & Class 3           & Class 4           & Class 5           & Class 6           & Class 7           & Class 8           & Class 9           \\ \midrule
GANomaly \cite{ganomaly}    & 0.881             & 0.675             & 0.953             & 0.801             & 0.827             & 0.864             & 0.849             & 0.682             & 0.856             & 0.558             \\
AnoGAN \cite{anogan}      & 0.623             & 0.31              & 0.521             & 0.458             & 0.442             & 0.431             & 0.492             & 0.401             & 0.392             & 0.368             \\
EGBAD \cite{egbad}       & 0.783             & 0.294             & 0.523             & 0.506             & 0.453             & 0.436             & 0.593             & 0.398             & 0.523             & 0.358             \\
DenseNet-169 & \textbf{0.998265} & 0.994258          & \textbf{0.984126} & 0.980750          & 0.983918          & 0.992295          & \textbf{0.984011} & 0.997476          & 0.991551          & \textbf{0.999386} \\
ResNet-152   & 0.998050          & 0.994176          & 0.982025          & \textbf{0.981253} & 0.984338          & 0.989994          & 0.980970          & \textbf{0.998940} & 0.989815          & 0.998982          \\
Inception-V4 & 0.997676          & \textbf{0.994609} & 0.983431          & 0.980548          & \textbf{0.984617} & \textbf{0.992676} & 0.983624          & 0.997108          & \textbf{0.994305} & 0.999080           \\
\bottomrule
\end{tabular}
\end{table*}

\begin{table}[!ht]
    \centering
    \caption{AUC results for Concrete Crack dataset}
    \label{tab:concretecrack}
    \begin{tabular}{ll}
        \toprule
        Model        & AUC               \\ \midrule
        GANomaly \cite{ganomaly}    & 0.858             \\
        DenseNet-169 & \textbf{0.999998} \\
        ResNet-152   & 0.999986          \\
        Inception-V4 & 0.998462  \\ \bottomrule
    \end{tabular}
\end{table}

\begin{table}[!ht]
    \centering
    \caption{Confusion Matrix for Concrete Crack dataset using DenseNet-169}
    \label{tab:cmatrixdensenet}
    \begin{tabular}{l|l|c|c|c}
    \multicolumn{2}{c}{}&\multicolumn{2}{c}{Predicted}&\\
    \cline{3-4}
    \multicolumn{2}{c|}{}&Crack&No Crack&\multicolumn{1}{c}{Total}\\
    \cline{2-4}
    \multirow{2}{*}{Actual}& Crack & $3993$ & $7$ & $4000$\\
    \cline{2-4}
    & No Crack & $0$ & $4000$ & $4000$\\
    \cline{2-4}
    \multicolumn{1}{c}{} & \multicolumn{1}{c}{Total} & \multicolumn{1}{c}{$3993$} & \multicolumn{    1}{c}{$4007$} & \multicolumn{1}{c}{$8000$}\\
    \end{tabular}
\end{table}

\begin{table}[!ht]
    \centering
    \caption{Confusion Matrix for Concrete Crack dataset using ResNet-152}
    \label{tab:cmatrixresnet}
    \begin{tabular}{l|l|c|c|c}
    \multicolumn{2}{c}{}&\multicolumn{2}{c}{Predicted}&\\
    \cline{3-4}
    \multicolumn{2}{c|}{}&Crack&No Crack&\multicolumn{1}{c}{Total}\\
    \cline{2-4}
    \multirow{2}{*}{Actual}& Crack & $3975$ & $25$ & $4000$\\
    \cline{2-4}
    & No Crack & $1$ & $3999$ & $4000$\\
    \cline{2-4}
    \multicolumn{1}{c}{} & \multicolumn{1}{c}{Total} & \multicolumn{1}{c}{$3976$} & \multicolumn{    1}{c}{$4024$} & \multicolumn{1}{c}{$8000$}\\
    \end{tabular}
\end{table}

\begin{table}[!ht]
    \centering
    \caption{Confusion Matrix for Concrete Crack dataset using Inception-V4}
    \label{tab:cmatrixinceptionv4}
    \begin{tabular}{l|l|c|c|c}
    \multicolumn{2}{c}{}&\multicolumn{2}{c}{Predicted}&\\
    \cline{3-4}
    \multicolumn{2}{c|}{}&Crack&No Crack&\multicolumn{1}{c}{Total}\\
    \cline{2-4}
    \multirow{2}{*}{Actual}& Crack & $3952$ & $48$ & $4000$\\
    \cline{2-4}
    & No Crack & $51$ & $3949$ & $4000$\\
    \cline{2-4}
    \multicolumn{1}{c}{} & \multicolumn{1}{c}{Total} & \multicolumn{1}{c}{$4003$} & \multicolumn{1}{c}{$3997$} & \multicolumn{1}{c}{$8000$}\\
    \end{tabular}
\end{table}

\section{Conclusion}
This paper introduces an approach that applies network-based deep transfer learning to the task of anomaly detection by treating it as a supervised binary classification problem. The quantitative results obtained on different datasets including a real world dataset show that the proposed technique achieves state-of-the-art results in comparison to existing techniques. The proposed method also addresses one of the most important challenges of anomaly detection which is the  low availability of anomalous samples. This is demonstrated by the fact that the architectures were trained on low training samples per class and still the proposed method is able to separate the normal and abnormal classes competently. Future research can be conducted on the effect of source architecture choice, source dataset choice, effect of freezing first $n$ layers of the chosen architecture while training and the application of the method to other types of anomaly detection tasks. 
% conference papers do not normally have an appendix

% use section* for acknowledgement
%\section*{Acknowledgment}

% trigger a \newpage just before the given reference
% number - used to balance the columns on the last page
% adjust value as needed - may need to be readjusted if
% the document is modified later
%\IEEEtriggeratref{8}
% The "triggered" command can be changed if desired:
%\IEEEtriggercmd{\enlargethispage{-5in}}

% references section

% can use a bibliography generated by BibTeX as a .bbl file
% BibTeX documentation can be easily obtained at:
% http://www.ctan.org/tex-archive/biblio/bibtex/contrib/doc/
% The IEEEtran BibTeX style support page is at:
% http://www.michaelshell.org/tex/ieeetran/bibtex/
\bibliographystyle{IEEEtran}
% argument is your BibTeX string definitions and bibliography database(s)
\bibliography{IEEEabrv,refs}
%
% <OR> manually copy in the resultant .bbl file
% set second argument of \begin to the number of references
% (used to reserve space for the reference number labels box)

%\begin{thebibliography}{1}

%\bibitem{IEEEhowto:kopka}
%H.~Kopka and P.~W. Daly, \emph{A Guide to \LaTeX}, 3rd~ed.\hskip 1em plus
%  0.5em minus 0.4em\relax Harlow, England: Addison-Wesley, 1999.

%\end{thebibliography}

% that's all folks
\end{document}